\documentclass[10pt,twocolumn,letterpaper]{article}

\usepackage{wacv}              

\usepackage{graphicx}
\usepackage{amsmath}
\usepackage{amssymb}
\usepackage{booktabs}
\usepackage{algpseudocode}
\usepackage{algorithm}
\usepackage{subcaption}
\usepackage{array}
\usepackage[table]{xcolor}  
\usepackage{enumitem}      
\usepackage{caption}       
\usepackage[misc]{ifsym}
\usepackage[pagebackref,breaklinks,colorlinks]{hyperref}

\usepackage[capitalize]{cleveref}
\crefname{section}{Sec.}{Secs.}
\Crefname{section}{Section}{Sections}
\Crefname{table}{Table}{Tables}
\crefname{table}{Tab.}{Tabs.}

\def\wacvPaperID{1754} 
\def\confName{WACV}
\def\confYear{2025}

\begin{document}

\title{LLaVA-SpaceSGG: Visual Instruct Tuning for Open-vocabulary Scene Graph Generation with Enhanced Spatial Relations}

\author{Mingjie Xu$^{1}$\textsuperscript{*}, Mengyang Wu$^{2}$\textsuperscript{*}, Yuzhi Zhao$^{3}$\textsuperscript{†}, Jason Chun Lok Li$^{4}$, Weifeng Ou$^{5}$ \\
{\tt\small parasolohalo@gmail.com, yzzhao2-c@my.cityu.edu.hk} 
\and
$^{1}$Independent Researcher \quad
$^{2}$The Chinese University of Hong Kong \quad
$^{3}$City University of Hong Kong \quad \\
$^{4}$The University of Hong Kong \quad
$^{5}$Dongguan University of Technology
}
\maketitle

\def\thefootnote{*}\footnotetext{Equal Contribution.}
\def\thefootnote{†}\footnotetext{Corresponding Author.}

\begin{abstract}
Scene Graph Generation (SGG) converts visual scenes into structured graph representations, providing deeper scene understanding for complex vision tasks. However, existing SGG models often overlook essential spatial relationships and struggle with generalization in open-vocabulary contexts. To address these limitations, we propose LLaVA-SpaceSGG, a multimodal large language model (MLLM) designed for open-vocabulary SGG with enhanced spatial relation modeling. To train it, we collect the SGG instruction-tuning dataset, named SpaceSGG. This dataset is constructed by combining publicly available datasets and synthesizing data using open-source models within our data construction pipeline. It combines object locations, object relations, and depth information, resulting in three data formats: spatial SGG description, question-answering, and conversation. To enhance the transfer of MLLMs' inherent capabilities to the SGG task, we introduce a two-stage training paradigm. Experiments show that LLaVA-SpaceSGG outperforms other open-vocabulary SGG methods, boosting recall by 8.6\% and mean recall by 28.4\% compared to the baseline. Our codebase, dataset, and trained models are publicly accessible on GitHub at the following URL: \url{https://github.com/Endlinc/LLaVA-SpaceSGG}.

\end{abstract}

\section{Introduction}
\label{sec:intro}
\begin{figure*}[t]
    \centering
    \includegraphics[width=\linewidth]{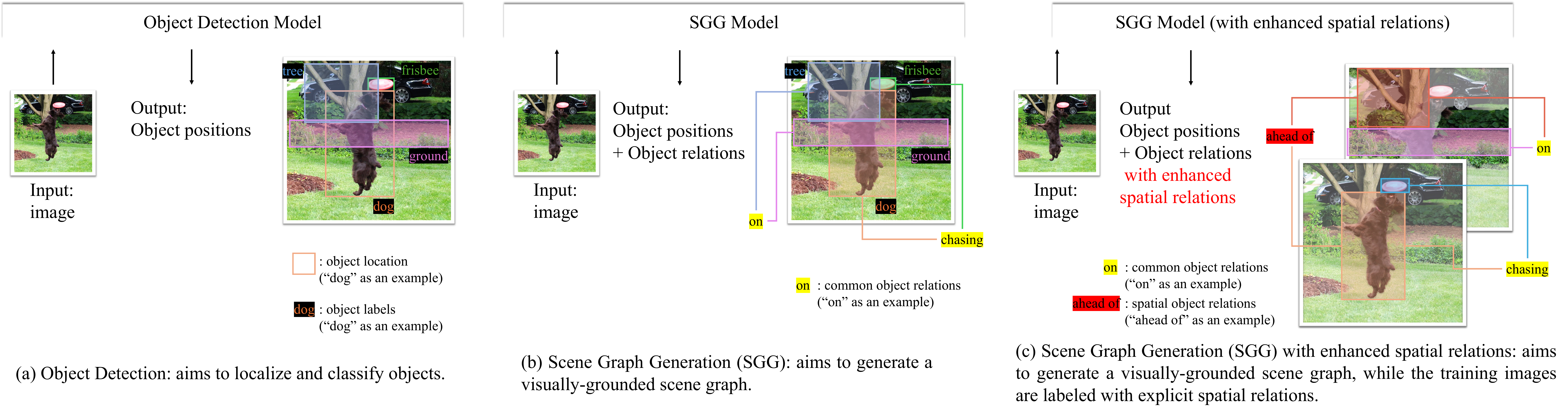}
    \caption{The illustration of different tasks: (a) Object Detection, (b) Scene Graph Generation (SGG), and (c) Scene Graph Generation (SGG) with enhanced spatial relations. By additionally leveraging spatial relationships, we propose the LLaVA-SpaceSGG framework.}
    \label{fig:diff}
\end{figure*}

Scene Graph Generation (SGG) is a fundamental scene understanding task that involves detecting the entities and predicting their relationships in an image to form a scene graph (see Figure \ref{fig:diff} (a) and (b)). The scene graph can be formulated as several text tuples of (subject, predicate, object), where the nodes denote objects and the edges denote relationships between different object pairs, respectively. Since the scene graph is a concise semantic representation of an image, it can be an intermediate feature for complex vision tasks. For instance, it has been applied in diverse downstream tasks such as visual question answering \cite{antol2015vqa, goyal2017making, li2023stablellava, chen2024internvl}, image captioning \cite{lin2014microsoft, li2017scene, sharma2018conceptual, yu2022coca, chen2023sharegpt4v}, image retrieval \cite{johnson2015image, schroeder2020structured, zeng2021conceptual}, etc.

Recent approaches have attempted to generate scene graphs in a supervised manner, yielding remarkable results. Nonetheless, we have identified two challenges that constrain the overall performance:

1) Open-vocabulary SGG: Existing SGG methods often require direct supervision with a fixed set of labels and their generalization ability on open-set images is unsatisfactory;

2) Lack of spatial relations: Since existing SGG datasets are mainly annotated on 2D images, the annotation progress mainly focuses on common relationships and neglects 3D spatial relationships between certain objects.

In pursuit of open-vocabulary SGG, recent approaches, exemplified by ASMv2 \cite{wang2024all}, have integrated state-of-the-art vision-language models like CLIP \cite{radford2021learning} and LLaVA \cite{liu2023visual}, leveraging rich and diverse training data encompassing various modalities. Nonetheless, these methods tend to overlook the crucial 3D spatial relationships that form integral elements of SGG. To emphasize spatial relations, Pu \etal\cite{pu2023spatial} and Li \etal\cite{li2018factorizable} integrated spatially specific blocks to assimilate spatial correlations and enhance spatial contextual understanding. However, it does not efficiently balance the original information and new information extracted by proposed blocks.


To address the two challenges simultaneously, we propose LLaVA-SpaceSGG, specifically designed to tackle both open-vocabulary SGG and spatial relationship modeling (see Figure \ref{fig:diff} (c)). We have extended the LLaVA-1.5 framework \cite{lu2016visual,liu2024improved} and curated a SpaceSGG instruction-tuning dataset. Then, we introduce a two-stage training paradigm to train the LLaVA-SpaceSGG. In the first stage, we align an image model (e.g., CLIP\cite{radford2021learning}) with a text model (e.g., Llama 2\cite{touvron2023llama}), enabling the model to excel in open-vocabulary SGG, leveraging the vast pre-training datasets. In the second stage, we refine the model's comprehension of region-level spatial relationships, crucial for SGG. This dual-phase approach comprises a pre-training stage succeeded by an instruction-tuning phase, akin to ASMv2 \cite{wang2024all}. To further enhance spatial understanding, we fully exploit SGG-related data in the second instruction-tuning phase, incorporating both the general SGG instruction-tuning dataset from \cite{wang2024all} and our newly created SpaceSGG dataset.

Our dataset introduces two key improvements. Firstly, it captures both plane and depth coordinates, enriching the spatial relationships (such as front-back relationships) between objects. Specifically, we first use a depth estimation algorithm \cite{yang2024depth} to generate a depth map from an image, then construct a 3D scene by\cite{Duzceker2021DeepVideo}, and finally extract 3D SGG from the 3D scene. Secondly, our dataset generates three distinct data formats: spatial descriptions (SpaceSGG-Desc), single-turn question answering (SpaceSGG-QA), and multi-turn conversations (SpaceSGG-Conv) to enhance the model’s spatial reasoning capabilities. SpaceSGG-Desc includes both plane and depth SGG descriptions. SpaceSGG-QA emphasizes the spatial relationships between two objects by depth comparison and multi-view questions. SpaceSGG-Conv contains the complete reasoning process from an image to SGG based on chain-of-thought (CoT) multi-turn dialogue \cite{wei2022chain}.

To evaluate the ability of the proposed LLaVA-SpaceSGG, we conduct experiments on a general Panoptic Scene Graph dataset (PSG) \cite{yang2022panoptic}. To further examine the spatial understanding ability, we construct a spatial relation validation dataset, which contains 271 labeled question-answer pairs on the COCO dataset \cite{lin2014microsoft}. Our LLaVA-SpaceSGG outperforms current state-of-the-art models by 8.6\% recall and by 28.4\% mean recall in the PSG validation set. It also outperforms existing methods with respect to an accuracy of 3.8\% in the proposed spatial relation validation set. Experiments show that LLaVA-SpaceSGG is able to discover, map, and predict richer spatial relationships while others do not. And it demonstrates that the SpaceSGG dataset greatly contributes to improving the model’s understanding of spatial relationships.


In summary, there are three main contributions:

1) To enhance spatial understanding in SGG, we collect the SpaceSGG dataset, along with a novel data generation pipeline. This dataset integrates both 2D and 3D scene information, resulting in a more comprehensive representation of object relations that captures spatial context, object positions, and depth. This fusion addresses critical limitations in existing SGG datasets, which often lack detailed spatial information.

2) Utilizing the SpaceSGG dataset, we develop LLaVA-SpaceSGG, a multimodal large language model designed specifically for open-vocabulary SGG tasks. In order to enhance the adaptation of MLLMs to the SGG domain, we propose a task-specific two-stage training strategy. This methodology notably enhances the model's capacity to interpret spatial relationships within intricate visual contexts.

3) The LLaVA-SpaceSGG model showcases the state-of-the-art performance on the well-established PSG validation set, surpassing current methods in recall and mean recall. Furthermore, to evaluate the model's proficiency in abstracting spatial relationships, we present a novel spatial relation validation set. Our model attains reliable and consistent performance on this new benchmark, underscoring its efficacy in capturing spatial dynamics.


\begin{figure*}[t]
    \centering
    \includegraphics[width=1\linewidth]{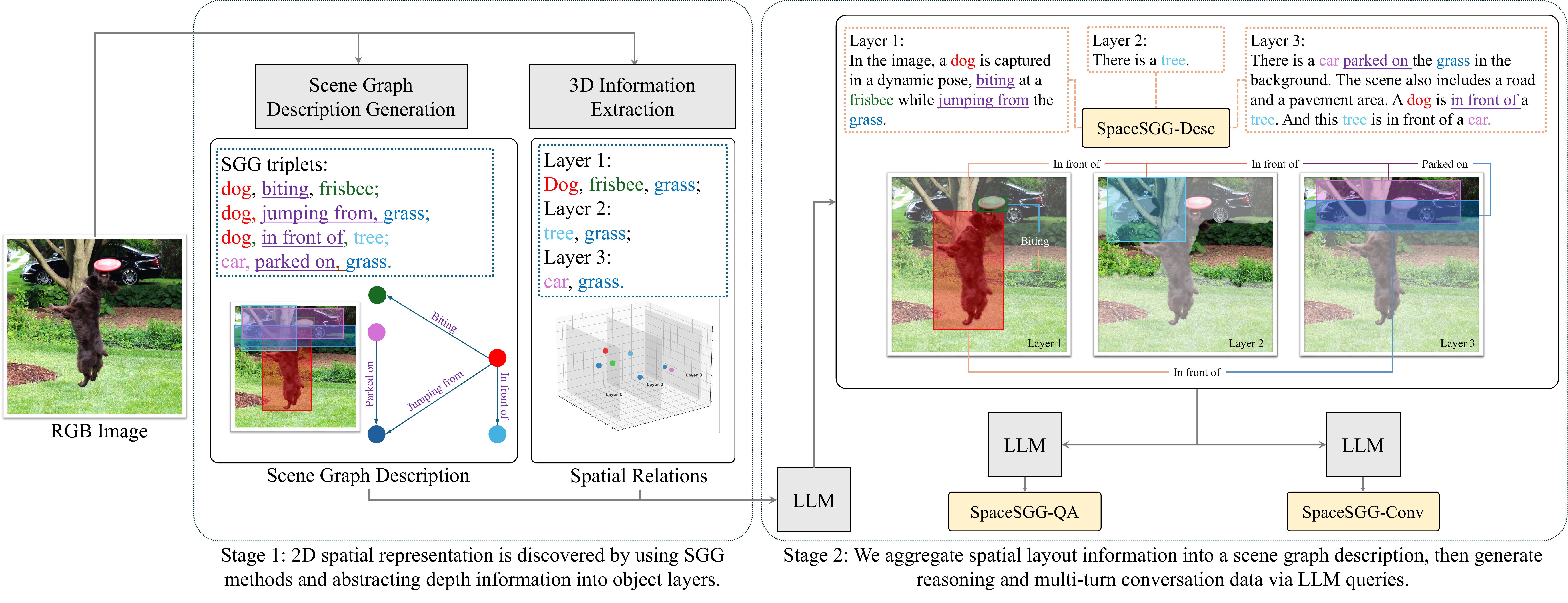}
    \caption{SpaceSGG dataset construction pipeline. We utilize both SGG description and spatial relationships, where we generate 3 types of data: spatial scene detailed descriptions (SpaceSGG-Desc), QA (SpaceSGG-QA), and multi-turn conversations (SpaceSGG-Conv).}
    \label{fig:overall-pipe}
\end{figure*}

\section{Related Work}
\subsection{Scene Graph Generation (SGG)}

SGG has become increasingly important in the computer vision area. The interest in this field was initially sparked by Lu \etal\cite{lu2016visual}, which focuses on the relationships between objects that can be abstractly represented by a graph of nodes and edges. Such scene graphs are immensely helpful for models to understand interactions among objects within images. However, early works\cite{yang2018graph,xu2017scene,li2017scene,chen2019knowledge} in this field overly simplified the scenes used for training, containing only a few objects and thus leading to an over-concentration of object relationships, which hindered the model's ability to learn generalized representational knowledge. Subsequently, Yang \etal\cite{yang2022panoptic} proposed conducting a comprehensive segmentation of the entire image and attempted to identify all relevant relationships between the resulting segments, thereby enriching the complexity of objects and their relationships. We aim to improve the performance of SGG by focusing on a fundamental yet underexplored aspect: spatial relationships. These relationships, which naturally exist between all objects, play a crucial role in understanding scenes. By addressing this gap, our method complements traditional approaches and significantly enhances model performance in SGG tasks.

\subsection{Depth Estimation and 3D Reconstruction}

Monocular depth estimation (MDE) has evolved significantly from early methods, which relied on handcrafted features and struggled with complex scenes due to their dependence on explicit depth cues \cite{hoiem2007recovering,ce2008sift}. The introduction of deep learning transformed MDE, with Eigen \etal\cite{eigen2014depth} pioneering a multi-scale fusion network for depth regression. Subsequent studies reframed regression as a classification task \cite{bhat2021adabins,li2024binsformer}, enhancing accuracy through improved priors and objective functions\cite{li2015depth,shao2023nddepth}.
For 3D reconstruction, classical methods computed dense depth per view \cite{schonberger2016pixelwise} and used techniques like Delaunay triangulation \cite{lee1980two} and Poisson surface reconstruction \cite{kazhdan2006poisson}. Recent deep learning approaches, such as ATLAS \cite{murez2020atlas}, NeuralRecon \cite{sun2021neuralrecon}, and TransformerFusion \cite{bozic2021transformerfusion}, bypass traditional depth estimation by backprojecting 2D features into 3D space, though they incur high computational costs. These advancements have significantly improved the reconstruction of spatial scenes and the generation of inter-object spatial relationships.

\subsection{Multimodal Large Language Models (MLLM)}

In recent years, significant breakthroughs have been made in visual scene understanding. Models trained on large-scale image-text pairs\cite{radford2021learning} have demonstrated powerful performance in various vision tasks. Researchers have further enhanced the model's performance\cite{li2021align,yu2022coca}, enabling Vision Language Models (VLM) to be applied in an expanding array of fields. Recently, the remarkable capabilities of Large Language Models (LLM) have led to a proliferation of LLM-based multimodal models\cite{brown2020language,chen2024internvl,liu2024visual,chen2024far,li2023blip,zhu2023minigpt}. These MLLM inherit robust understanding and reasoning abilities. Additionally, designing prompts\cite{brown2020language,wei2022chain,jia2022visual} enables MLLMs to reason about previously unseen tasks, and adding extra information\cite{chen2024spatialvlm,huang2024language,wang2024all} to inputs can enhance the MLLM's performance on tasks. We attempt to leverage this capability of MLLMs and have collected a dataset that can be utilized for these models, which we call the SpaceSGG dataset. This format is intended to enhance the model's understanding of both scenes and space, integrating text generation, object localization, relationship understanding, and spatial comprehension.

\section{Methodology}
\label{sec:method}

To equip the model with scene graph generation capabilities and enhance spatial relation recognition, we first introduce the data construction pipeline for building the SpaceSGG dataset, which integrates spatial and scene graph information. Building upon this, we then detail the training paradigm for LLaVA-SpaceSGG.

\subsection{SpaceSGG Dataset Construction}

We hypothesize that the model's inaccurate predictions of spatial relationships stem from a lack of relevant spatial information annotations and the inadequate integration of spatial and scene graph information. To address this, we design a two-stage data generation process, as shown in Figure \ref{fig:overall-pipe}. In the first stage, scene graph description (Section \ref{sec:sg-desc}) and spatial layout are extracted from the 2D image (Section \ref{sec:spa-rel}). In the second stage, the scene graph triplets and spatial layout are fused into a spatial scene graph description (Section \ref{sec:spa-desc}), followed by spatial QA and spatial multi-turn conversations (Section \ref{subsec:qagen}). In this paper, we adopt the Llama 3 70B \cite{llama3modelcard} as the data generator.

\subsubsection{Scene Graph Description Generation}
\label{sec:sg-desc}
Scene graph description provides a detailed description, which serves as an intermediate state for the model to further generate a layered, comprehensive description of the image. To achieve this goal, we query GPT-4V\cite{achiam2023gpt} to generate responses that link the objects and predicates mentioned in the generated response to specific regions within the image, by following work\cite{wang2024all}.

\begin{figure}[t]
    \centering
    \includegraphics[width=0.98\linewidth]{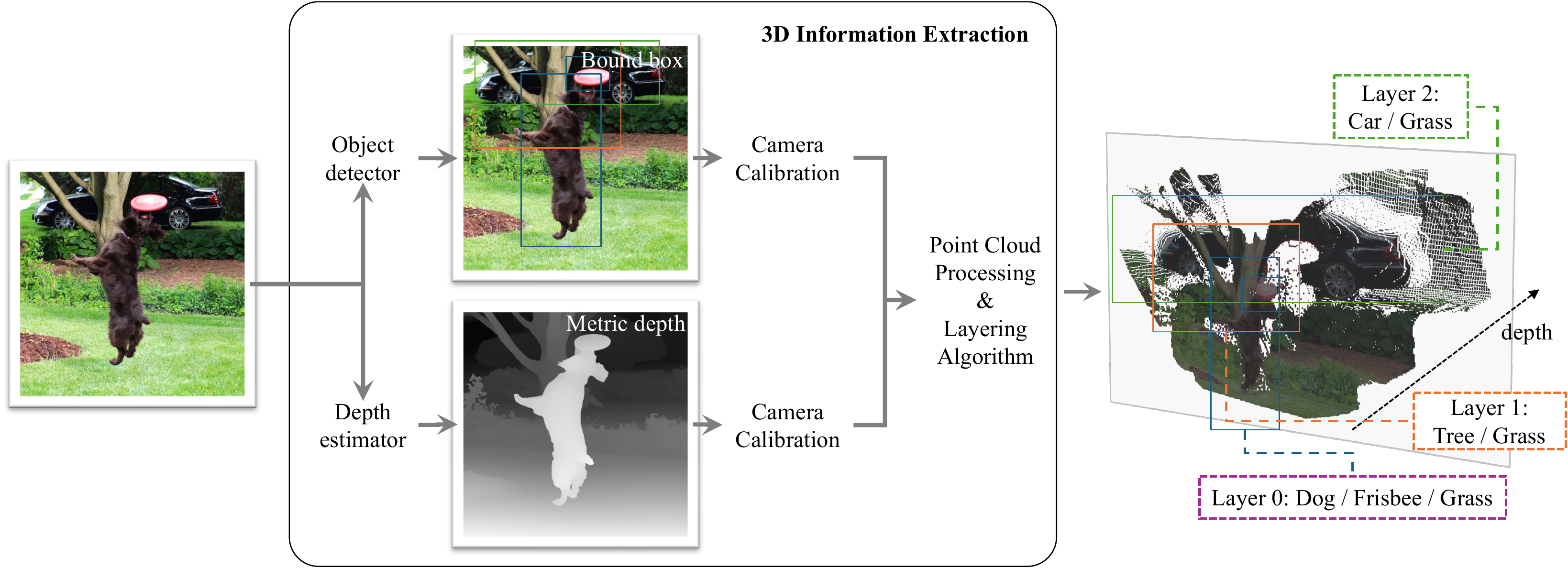}
    \caption{3D Information Extraction: We retrieve the spatial layering distribution of the input images with the assistance of object detectors and depth estimator.}
    \label{fig:enter-label}
\end{figure}

\subsubsection{Spatial Relations Extraction}
\label{sec:spa-rel}
To ensure that the spatial scene graph data contains more accurate spatial relationships, we need to reconstruct 3D spatial relationships from 2D images. We developed a pipeline that converts 2D images into depth maps and then into 3D point clouds. Firstly we apply the depth-anything~\cite{yang2024depth} model as our base depth detector and align the relative depth with each detected object (see Figure ~\ref{fig:enter-label}). 
Then, a camera calibration\cite{cheng2024spatialrgpt} has been appended including two parts: 1) estimating intrinsic parameters to convert depth maps into 3D point clouds, and 2) ensuring that scene relationships are consistent. This allows us to create a rotation matrix to convert the point cloud for each object. The point cloud data are saved accordingly with the objects for further conversion. 

\begin{algorithm}
    \caption{Relative Spatial Relation of Object A and B}
    \begin{algorithmic}
        \State \textbf{Input:} object A, object B, image
        \State depthmap $\gets$ DepthAnything(image)
        \State pointcloud $\gets$ CameraCalibration(depthmap)
        \State points A $\gets$ pointcloud[object A]
        \State points B $\gets$ pointcloud[object B]
        \State z\_range\_a $\gets$ [points A.Z\_min, points A.Z\_max]
        \State z\_range\_b $\gets$ 
        [points B.Z\_min, points B.Z\_max]
        \If{z\_range\_a.min $<$ z\_range\_b.min and z\_range\_a.max $>$ z\_range\_b.max}
                \State z\_range\_a \text{ covers } z\_range\_b
        \EndIf
        \State record the relative spatial relation of objects A and B.
    \end{algorithmic}
\label{algo:rel-relation}
\end{algorithm}

\begin{algorithm}
    \caption{Devide objects into layers}
    \begin{algorithmic}
        \State \textbf{Input:} objects, image
        \State Initialize layer\_list.
        \For{object A \textbf{in} objects}
            \State Initialize basic\_layer\_element\_flag with True.
            \For{object B \textbf{in} objects}
                \If{object A \text{ is overlapping with } object B}
                    \State basic\_layer\_element\_flag $\gets$ False
                \EndIf
            \EndFor
            \If{basic\_layer\_element\_flag}
                \State Add object A into layer\_list
            \EndIf
        \EndFor
        \State Sort objects in layer\_list with depth
        \For{object A \textbf{in} layer\_list}
            \State Initialize sub\_layer\_list of A
            \For{object B \textbf{in} objects}
                \If{object A \text{ is covered by } object B}
                    \State  Add object B into sub\_layer\_list of A
                \EndIf
            \EndFor
        \EndFor
        \State record the layer\_list and corresponding sub\_layer\_list.
    \end{algorithmic}
\label{algo:layering}
\end{algorithm}

\subsubsection{SpaceSGG Description Generation}
\label{sec:spa-desc}
\begin{figure}[t]
    \centering
    \includegraphics[width=0.9\linewidth]{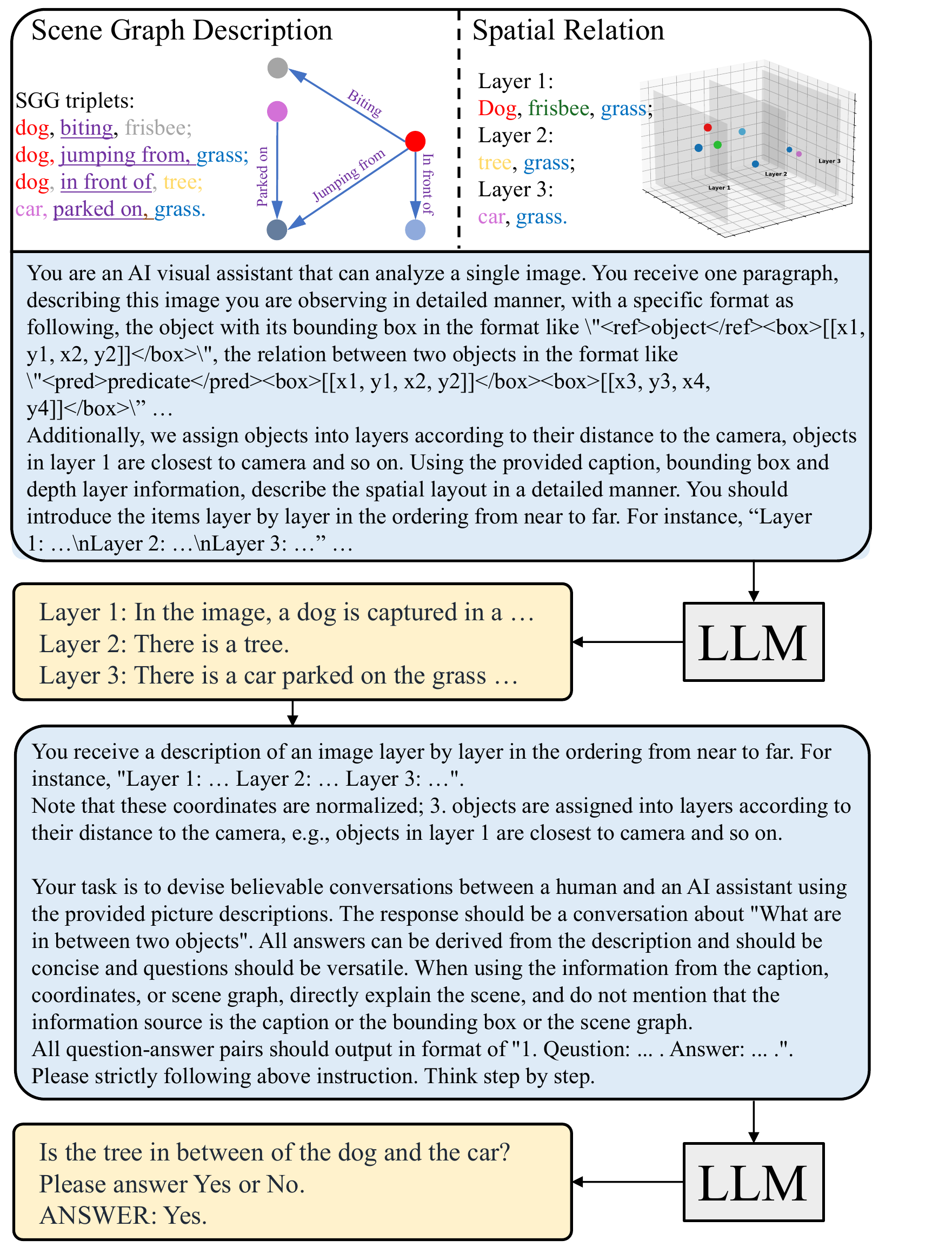}
    \caption{An example of SpaceSGG-Desc, SpaceSGG-QA, and SpaceSGG-Conv generation process.}
    \label{fig:fuse-data}
\end{figure}

After stage 1, we focus on naturally integrating spatial layout and scene graph information in stage 2.
%
We define explicit spatial layering to cluster and organize objects in space using the following algorithms (see Algorithm \ref{algo:rel-relation} and Algorithm \ref{algo:layering}):
1) one object with a depth range (by retrieving the minimum and maximum z-axis value) enclosed inside other objects' depth range is considered covered by others; 
2) objects not covering any other objects are considered to be the basic objects representing an individual layer; 
3) we check all the objects and select all the basic objects as the first elements in each layer;
4) we check all the other objects and assign them to different layers, such that larger objects may be added to multiple layers.
%
%
This definition emphasizes the spatial information of objects for further processing.
Then, we use a large language model \cite{llama3modelcard} to reorganize the language from the previous step, emphasizing the spatial position information of objects and expressing the hierarchical information of objects in space. More specifically, we prepare the structural scene graph description (See Figure ~\ref{fig:fuse-data} top left) and the spatial relations after spatial layering (See Figure ~\ref{fig:fuse-data} top right) as input, and apply well-designed instructions (See Figure ~\ref{fig:fuse-data} blue regions) for prompting LLMs. The final output is grouped by layers termed \textbf{SpaceSGG-Desc}.

\subsubsection{SpaceSGG QA and Conversation Generation}\label{subsec:qagen}
To enhance the model's spatial understanding and scene graph generation, we generate spatial question-answer pairs from scene graph descriptions, including single-turn QA (\textbf{SpaceSGG-QA}) and multi-turn conversation data (\textbf{SpaceSGG-Conv}) using LLM queries. Prompts (see Figure \ref{fig:fuse-data} bottom) are designed to split scene graph descriptions, triplets, and object depth distributions into questions about front-back judgment, up-down judgment, multi-object sorting, occlusion, and more (see Supplementary Materials A for examples). The SpaceSGG dataset combines spatial descriptions (SpaceSGG-Desc), single-turn QA (SpaceSGG-QA), and multi-turn conversations (SpaceSGG-Conv), encompassing 20K diverse scenes and their spatial structures.


\begin{figure}[t]
    \centering
    \includegraphics[width=1\linewidth]{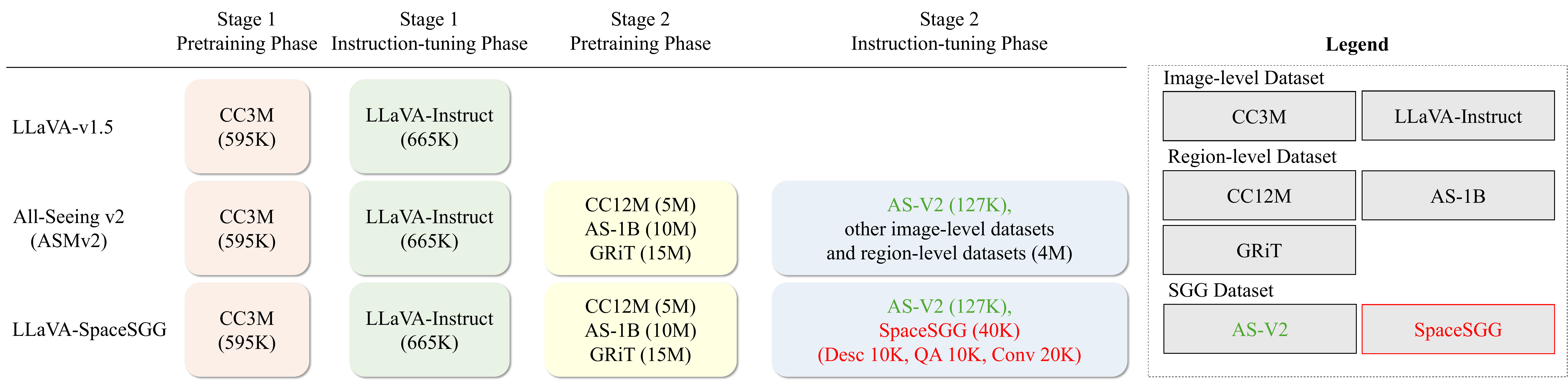}
    \caption{Our proposed training paradigm and used training dataset.}
    \label{fig:paradigm}
\end{figure}

\subsection{Training Paradigm}
To enhance spatial SGG knowledge transfer to MLLMs, we propose a specialized training paradigm inspired by LLaVA-1.5~\cite{liu2024visual} and ASMv2~\cite{wang2024all}. As shown in Figure~\ref{fig:paradigm}, our approach consists of two stages, each with a pre-training and instruction-tuning phase. In the first stage, we align modality features using image-level datasets, following the LLaVA-1.5~\cite{liu2024visual} setup. In the second stage, pre-training uses region-level datasets to refine feature discovery and grounding, while instruction-tuning combines our proposed dataset with existing SGG data to enhance the model's understanding of scene relationships and spatial layouts. The datasets used in each stage are detailed below.


In the first stage, we use image-level datasets CC3M~\cite{sharma2018conceptual} and LLaVA-Instruct~\cite{liu2024visual}. For the second stage, we use region-level datasets CC12M~\cite{sharma2018conceptual}, AS-1B~\cite{wang2023all}, and GRiT~\cite{peng2023kosmos} to enhance the model's ability to discover subtle features and improve grounding. Unlike ASMv2, which incorporates 4M images from additional datasets like OCR-VQA~\cite{ocrvqa} and TextVQA~\cite{textvqa}, our instruction-tuning phase focuses solely on SGG datasets, including AS-V2~\cite{wang2024all} and the proposed SpaceSGG dataset.


\section{Experiments}

We conduct both quantitative and qualitative analysis with state-of-the-art methods on the public PSG dataset \cite{yang2022panoptic} and the proposed spatial relation validation set. Then, we design several ablation study settings to show the effectiveness of our training paradigm and SpaceSGG dataset.

\subsection{Validation Sets and Metrics}

\textbf{PSG Validation Set.} We use the PSG dataset \cite{yang2022panoptic} to evaluate the open-vocabulary SGG capabilities of existing models. The dataset includes 49K images and 56 relationships, six of which are positional (e.g., over, in front of, beside, on, in, attached to). We evaluate both closed-set methods \cite{xu2017scene,zellers2018neural,tang2019learning,lin2020gps,yang2022panoptic} and open-set methods \cite{zhao2023textpsg,wang2024all}.


\textbf{Spatial Relation Validation Set.} To assess spatial understanding, we use the proposed spatial relation validation set. We randomly select 30 images from COCO-Val-2017 and generate two types of questions (QA and multi-turn conversations) using the data generation pipeline in Section \ref{subsec:qagen}. These are manually annotated as single-choice QA with factual corrections, resulting in 271 questions. This benchmark evaluates the model's spatial understanding of scenes.


\textbf{Metrics}. Following \cite{yang2022panoptic,zhao2023textpsg,wang2024all}, we report triplet Recall and mean Recall (mRecall) for each predicate category in the open-vocabulary SGG task. A scene graph consists of triplets (subject, predicate, object), and a triplet is considered correct if the phrase labels are accurate and the subject and object locations match the ground truth with an Intersection over Union (IoU) greater than 0.5. Recall and mRecall are then computed as follows:


\begin{equation}
    \text{Recall} = \frac{\text{Number of predicates}}{\text{Total number of ground truth relationships}}.
\end{equation}
\begin{equation}
\text{mRecall} = \frac{1}{N} \sum_{i=1}^{N} \text{Recall}\ (i\in relation\ classes).
\end{equation}


\begin{figure*}
    \centering
    \includegraphics[width=0.99\linewidth]{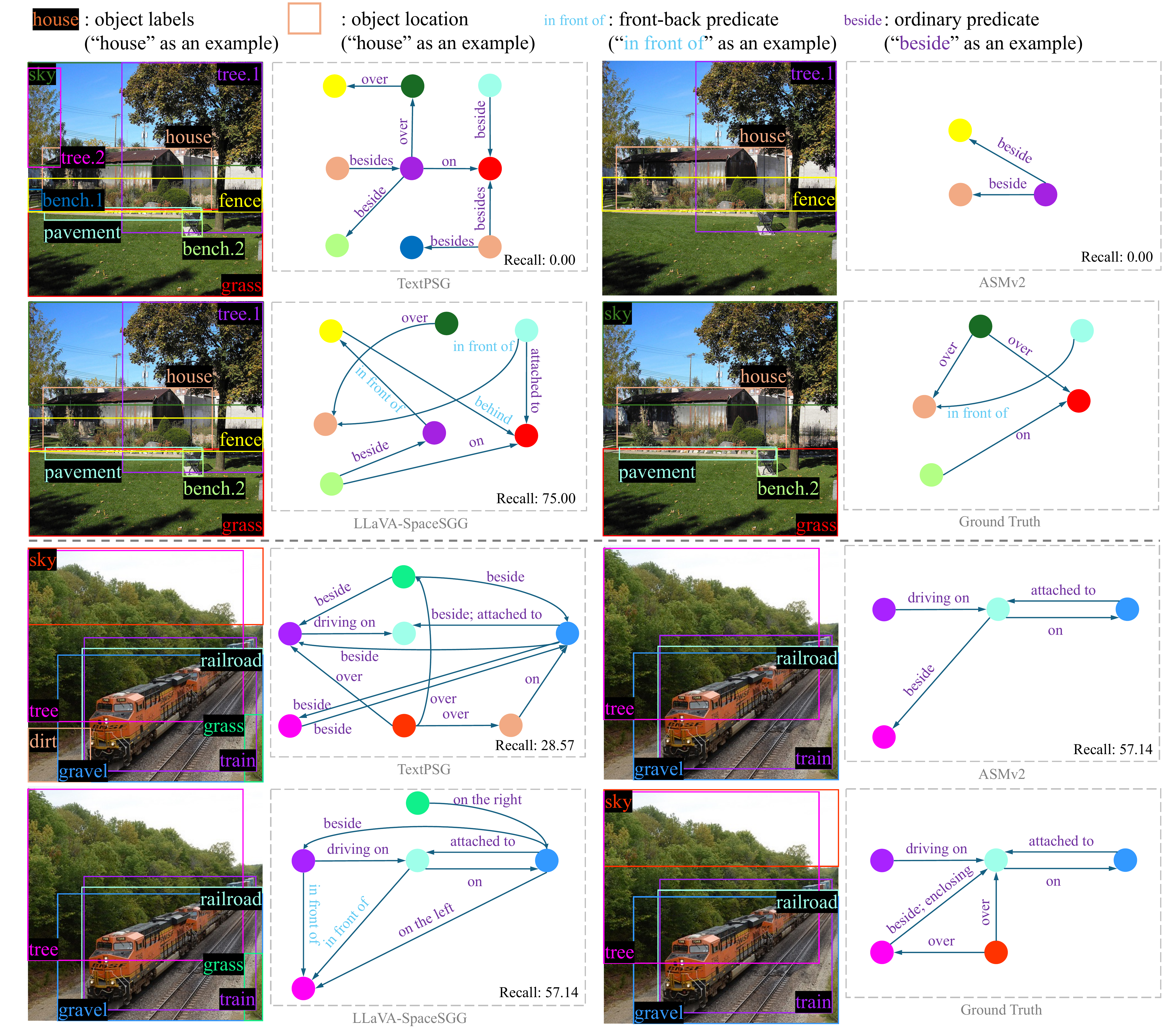}
    \caption{Qualitative result of open-vocabulary SGG, particularly from traditionally state-of-arts models. Note that the cyan-coloured predicate denotes a precise front-back relationship.}
    \label{fig:qualitative-result}
\end{figure*}

\subsection{Comparison with SoTA Methods}

To test the effectiveness of our SpaceSGG dataset and training method, we compared LLaVA-SpaceSGG with state-of-the-art models\cite{wang2024all,zhao2023textpsg, xu2017scene,zellers2018neural,tang2019learning,lin2020gps,yang2022panoptic} on the PSG \cite{yang2022panoptic} dataset and our proposed spatial relation validation set.

\begin{table}[t]
    \centering
    \begin{tabular}{lcc}
        \toprule
        Model               & Recall & mRecall \\
        \midrule
        \textit{Close-ended SGG}\\
        IMP       & 16.5   & 6.5     \\
        MOTIFS    & 20.0   & 9.1     \\
        VCTree    & 20.6   & 9.7     \\
        GPSNet    & 17.8   & 7.0     \\
        PSGFormer & 18.6   & 16.7    \\
        \hline
        \textit{Open-ended SGG}\\
        TextPSG  & 4.8    & --      \\
        ASMv2        & 14.2   & 10.3    \\
        LLaVA-SpaceSGG    & \textcolor{red}{15.43} & \textcolor{red}{13.23} \\
        \bottomrule
    \end{tabular}
    \caption{Open-vocabulary SGG performance comparison between our model and other specialist models. The \textcolor{red}{red} denotes the best results across all methods.}
    \label{tab:model_comparison}
\end{table}

\subsubsection{PSG Validation Set}

We compare our model on the PSG validation set under the open-vocabulary SGG setting against open-set models \cite{wang2024all,zhao2023textpsg} and closed-set models \cite{xu2017scene,zellers2018neural,tang2019learning,lin2020gps,yang2022panoptic}. As shown in Table \ref{tab:model_comparison}, our model achieves state-of-the-art performance, outperforming ASMv2 by 8.6\% in Recall and 28.4\% in mRecall. Against closed-set models, it demonstrates strong performance with a recall of 15.43 and a mean recall of 13.23.

As shown in Figure \ref{fig:qualitative-result} (bottom example), TextPSG often generates redundant relationships. However, our model produces concise scene graphs with accurate spatial relations (e.g., ``in front of,'' ``inside of,'' ``beside''), leading to higher recall. Compared to ASMv2, our model captures more nuanced object relationships, enriching scene graphs and improving scene understanding. Additional results are provided in Supplementary Materials B.



\begin{table}[t]
    \centering
    \begin{tabular}{lc} 
        \toprule
        Model & Accuracy (\%) \\
        \midrule
        Random Choice & 25.00 \\
        LLaVA-1.5-13B & 45.13 \\
        ASMv2-13B & 50.52 \\
        LLaVA-SpaceSGG & \textcolor{red}{52.48} \\
        \bottomrule
    \end{tabular}
    \caption{Comparison of model accuracies for spatial understanding tasks. Our model outperforms established benchmarks.}
    \label{tab:model_space_accuracy}
\end{table}

\subsubsection{Spatial Relation Validation Set}

To validate the spatial capabilities of LLaVA-SpaceSGG, we compare its spatial relationship prediction accuracy with state-of-the-art open-source MLLMs \cite{liu2024improved,wang2024all} on our spatial relation validation set. As shown in Table \ref{tab:model_space_accuracy}, our model outperforms existing methods, with ASMv2-13B \cite{wang2023all} being the closest competitor due to its use of scene graph data, though it struggles with 3D spatial contexts.

Figure \ref{fig:qualitative-result} (top example) illustrates that, compared to TextPSG and ASMv2, our method captures more detailed spatial relationships, such as front-back orientations (blue) and distinctions between terms like ``on the left,'' ``beside,'' ``attached to,'' ``on,'' and ``over'' (purple). This demonstrates the value of integrating high-quality spatial and scene graph data, which current state-of-the-art models lack.

\begin{figure}[t]
    \centering
    \includegraphics[width=\linewidth]{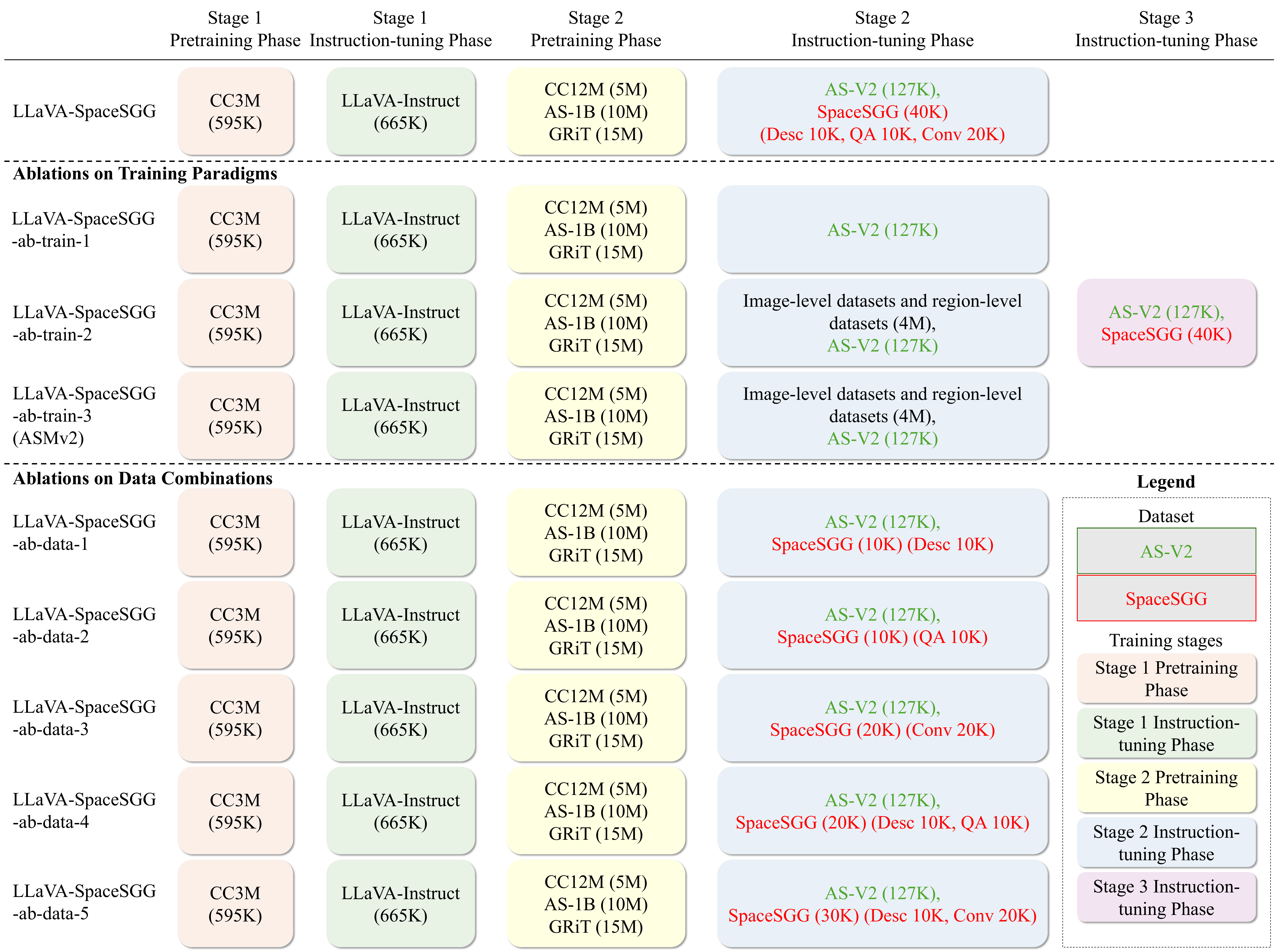}
    \caption{We conduct two types of ablation studies: training paradigm and data combination.}
    \label{fig:ablations}
\end{figure}

\subsection{Ablation Studies}

In this section, we conduct ablation studies to prove the effectiveness of our training paradigm and our proposed SpaceSGG dataset. All ablation settings can be found in Figure \ref{fig:ablations}. We first compare the effectiveness of the training paradigm in three settings (see Figure \ref{fig:ablations} ablations on training paradigms). Then, we conduct five ablation studies to validate the effectiveness of our proposed SpaceSGG dataset (see Figure \ref{fig:ablations} ablations on data combinations).

\subsubsection{Comparison on Different Training Paradigms}

We conduct experiments under three training paradigms to evaluate the interaction of our SpaceSGG dataset with other data, as shown in Table \ref{tab:training_evaluation}. The LLaVA-SpaceSGG-ab-train-1 setting performs worse than LLaVA-SpaceSGG, with a 6\% drop in recall, 21.9\% in mean recall, and reduced spatial validation accuracy, highlighting the benefits of mixing SpaceSGG with the AS-V2 dataset during stage 2 instruction tuning. LLaVA-SpaceSGG-ab-train-2, based on an additional SFT phase, also underperforms, demonstrating the efficiency and effectiveness of our two-stage training paradigm. Lastly, LLaVA-SpaceSGG-ab-train-3, which uses the existing ASMv2 training approach not tailored for SGG, achieves lower performance, confirming the superiority of our training design for SGG tasks.


\begin{table}[t]
    \centering
    \resizebox{\linewidth}{!}{
    \rowcolors{2}{gray!25}{white}  
    \begin{tabular}{p{3cm}ccc}  
        \toprule
        Ablation Setting & Recall & mRecall & Accuracy (\%) \\
        \midrule
        LLaVA-SpaceSGG\newline -ab-train-1 & 14.41 & 10.32 & 1.47 \\
        LLaVA-SpaceSGG\newline -ab-train-2 & 13.97 & 10.2  & 50.52 \\
        LLaVA-SpaceSGG\newline -ab-train-3 & 14.2  & 10.3  & 45.7  \\
        LLaVA-SpaceSGG & \textcolor{red}{15.43} & \textcolor{red}{13.23} & \textcolor{red}{52.48} \\
        \bottomrule
    \end{tabular}
    }
    \caption{We conducted ablation studies based on different training stages. For the specific training set-ups of various settings, please refer to Figure \ref{fig:ablations}.}
    \label{tab:training_evaluation}
\end{table}

\subsubsection{Comparison on Different Data Combinations}

To validate the effectiveness of the proposed SpaceSGG dataset, we conducted an ablation study by excluding specific terms from the SpaceSGG dataset to create five settings. First, LLaVA-SpaceSGG-ab-train-1, trained without the SpaceSGG dataset, shows reduced performance on the PSG validation set and a significant drop in the spatial benchmark, emphasizing the dataset's importance. Second, LLaVA-SpaceSGG-ab-data-1, LLaVA-SpaceSGG-ab-data-2, and LLaVA-SpaceSGG-ab-data-3 perform worse than the full model, as shown in Table \ref{tab:data_evaluation}, underscoring the value of data combination. Third, LLaVA-SpaceSGG-ab-data-4 and LLaVA-SpaceSGG-ab-data-5, which include SpaceSGG-QA and SpaceSGG-Conv alongside SpaceSGG-Desc, achieve higher spatial benchmark accuracy but lower Recall and mRecall than the baseline, which better balances these metrics.


\subsubsection{Comparison on Different Generative Models in Pipeline}

To evaluate the role of data generator in the proposed pipeline, we replaced the default Llama 3 70B \cite{llama3modelcard} with alternatives, including Qwen2.5 72B \cite{qwen2.5} and GPT-4o \cite{gpt4o}. The results are shown in Table \ref{tab:generative_evaluation}, which demonstrate that the data quality remains consistent across different generative models, with performance on Open Vocabulary SGG and the Spatial Validation set varying minimally compared to the final chosen model (i.e., Llama 3 70B). These findings suggest that the choice of generative model has a negligible impact on overall data quality.

\begin{table}[t]
    \centering
    \resizebox{\linewidth}{!}{ 
    \rowcolors{2}{gray!25}{white}  
    \begin{tabular}{p{3cm}ccc}  
        \toprule
        \textbf{Ablation Setting} & \textbf{Recall} & \textbf{mRecall} & \textbf{Accuracy (\%)} \\
        \midrule
        LLaVA-SpaceSGG\newline-ab-data-1 & \textcolor{blue}{14.86} & 10.92 & 12.74 \\
        LLaVA-SpaceSGG\newline-ab-data-2 & \textcolor{green}{14.53} & 11.07 & \textcolor{green}{37.21} \\
        LLaVA-SpaceSGG\newline-ab-data-3 & 14.24 & \textcolor{blue}{12.27} & 4.41 \\
        LLaVA-SpaceSGG\newline-ab-data-4 & 14.39 & \textcolor{green}{11.26} & \textcolor{red}{53.39} \\
        LLaVA-SpaceSGG\newline-ab-data-5 & 14.5 & 10.2 & 24.03 \\
        LLaVA-SpaceSGG & \textcolor{red}{15.43} & \textcolor{red}{13.23} & \textcolor{blue}{52.48} \\
        \bottomrule
    \end{tabular}
    }
    \caption{We experimented with different mixing ratios of our generated data. The \textcolor{red}{red}, \textcolor{blue}{blue}, and \textcolor{green}{green} colors denote the best, the second highest and the third highest results, respectively. For detailed experimental settings, please refer to Figure \ref{fig:ablations}.}
    \label{tab:data_evaluation}
\end{table}

\begin{table}[t]
    \centering
    \resizebox{\linewidth}{!}{ 
    \rowcolors{2}{gray!25}{white}  
    \begin{tabular}{p{3cm}ccc}  
        \toprule
        \textbf{Ablation Setting} & \textbf{Recall} & \textbf{mRecall} & \textbf{Accuracy (\%)} \\
        \midrule
        LLaVA-SpaceSGG\newline-ab-Qwen2.5& 14.22& 9.53& 51.68\\
        LLaVA-SpaceSGG\newline-ab-GPT-4o& 13.99& 10.94& \textcolor{red}{53.725}\\
        LLaVA-SpaceSGG\newline(Llama3)& \textcolor{red}{15.43}& \textcolor{red}{13.23}& 52.48\\
    \end{tabular}
    }
    \caption{We experimented with different generated data by generative models.}
    \label{tab:generative_evaluation}
\end{table}

Overall, these results demonstrate that the spatial and SGG information in our dataset is highly effective and significantly enhances the model's performance on SGG tasks.

\section{Conclusions}

In this paper, we tackle the problem of open-vocabulary scene graph generation by enhancing the spatial relations. Since most of the existing datasets often overlook 3D relations in SGG, we propose a data generation pipeline that integrates both 2D and 3D scene information to obtain more comprehensive relations. It results in 10K spatial scene detailed descriptions, 20K question answers, and 20K multi-turn conversations. Built upon it, we present the LLaVA-SpaceSGG model. Specifically, we explore a task-specific training paradigm, which contains two stages that improve the model's ability to perceive spatial relations in complex visual scenes. Finally, we compare the proposed LLaVA-SpaceSGG on a well-known PSG dataset and our proposed spatial relation validation set. The experiment results show that LLaVA-SpaceSGG surpasses the current state-of-the-art models in the open-vocabulary SGG task, achieving an 8.6\% improvement in recall and a 28.4\% improvement in mRecall. It also performs better than other MLLMs in producing spatial relations. In the future, we aim to further enhance the model's visual understanding and reasoning capabilities by using high-quality and diverse annotations.

\clearpage
{\small
\bibliographystyle{ieee_fullname}
\bibliography{ref}
}

\clearpage
\appendix

The supplementary material contains:

1) more ablation studies testing effectiveness of the proposed dataset;

2) more examples about the SpaceSGG dataset including 3 components (SpaceSGG-Desc, SpaceSGG-QA and SpaceSGG-Conv);

3) more visual examples about our proposed LLaVA-SpaceSGG prediction compare with other models (TextPSG, ASMv2).

\section{More Ablation Studies}

\begin{figure}[t]
    \centering
    \includegraphics[width=\linewidth]{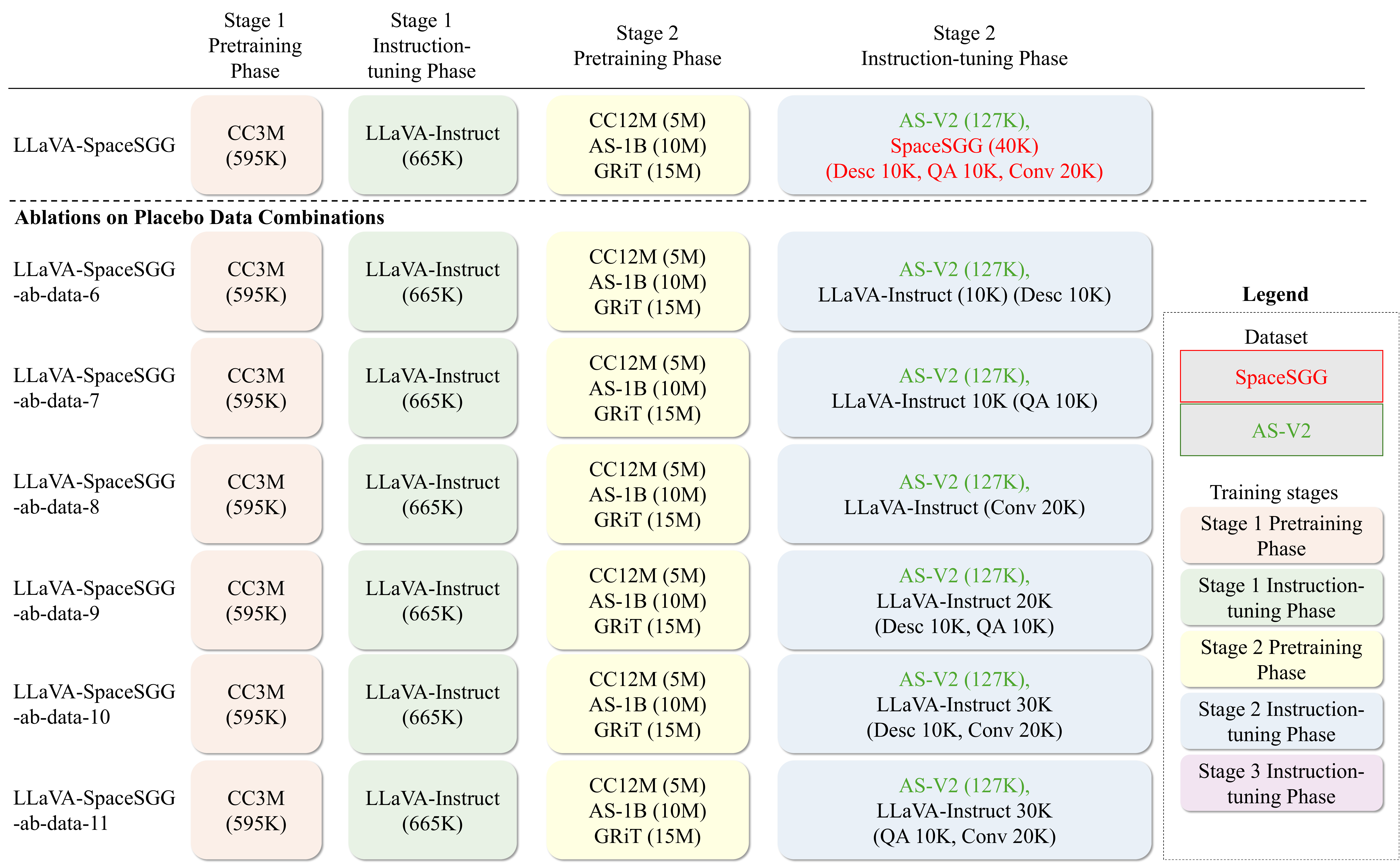}
    \caption{We conduct placebo ablation studies by testing the same data combination, replacing components in SpaceSGG with equivalent ones from the LLaVA-Instruct dataset.}
    \label{fig:placebo-ablations}
\end{figure}

To further validate the effectiveness of the proposed dataset, we replaced each element with equivalent components from the LLaVA-Instruct dataset, ensuring the same number of entries were sampled. The experimental settings are illustrated in Figure \ref{fig:placebo-ablations}. As shown in Table \ref{tab:placebo-evaluation}, these replacements did not improve the model's SGG performance or spatial understanding, further highlighting the significance of our dataset.

\section{SpaceSGG Dataset Examples}

\begin{figure*}
    \centering
    \includegraphics[width=0.85\linewidth]{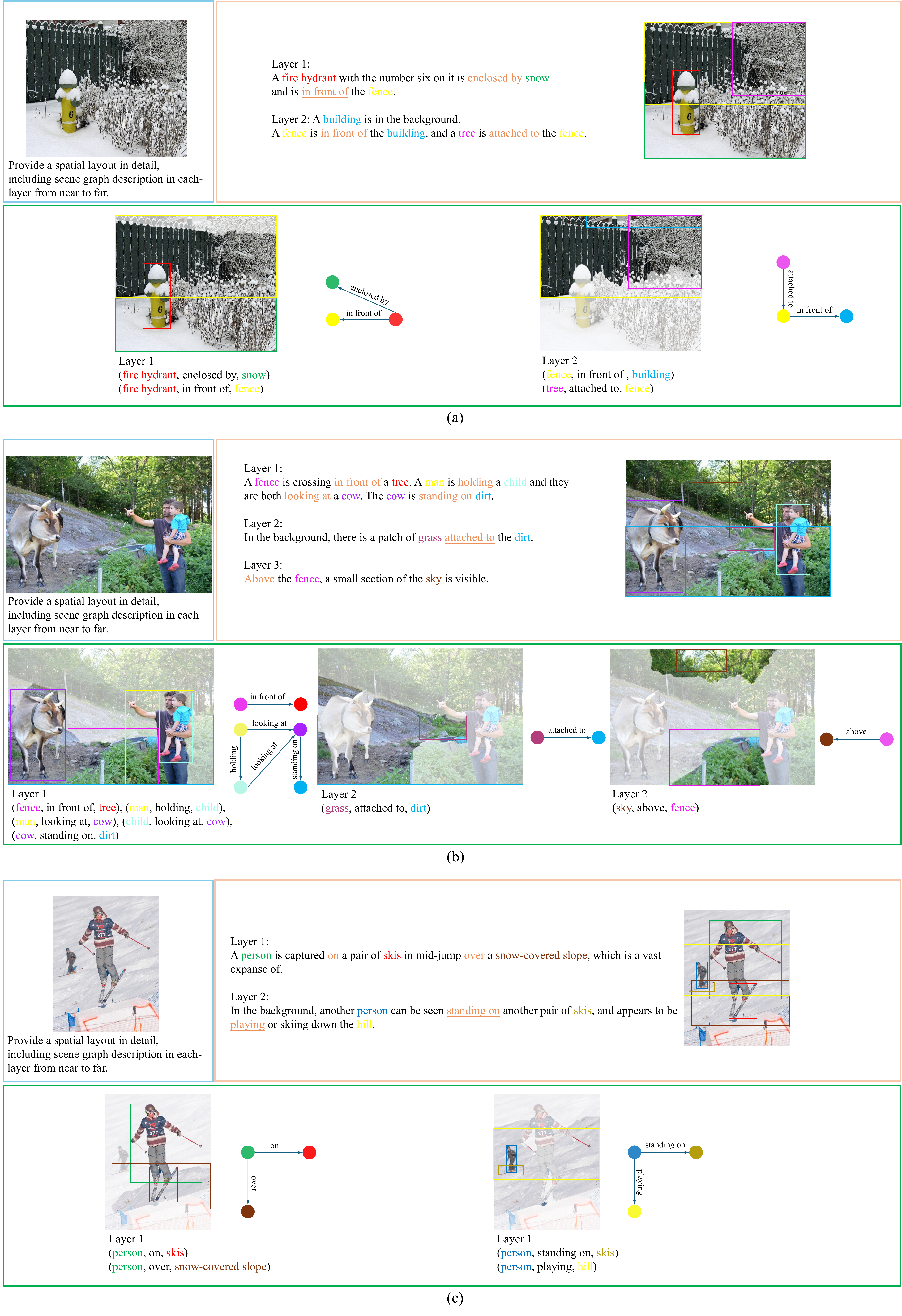}
    \caption{Data Examples of SpaceSGG-Desc in SpaceSGG.}
    \label{fig:supp-desc}
\end{figure*}

\begin{figure*}
    \centering
    \includegraphics[width=0.85\linewidth]{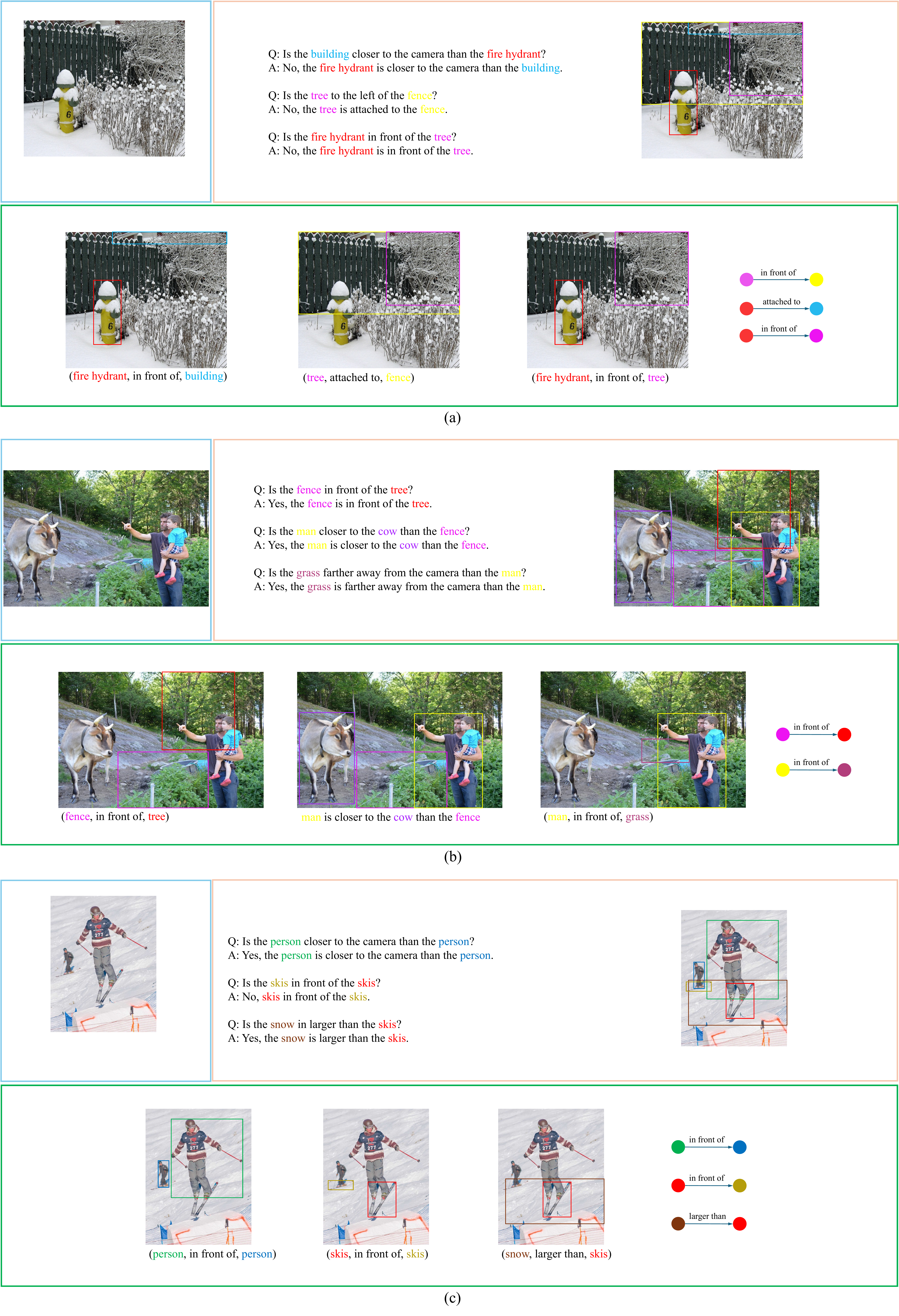}
    \caption{Data Examples of SpaceSGG-QA in SpaceSGG.}
    \label{fig:supp-qa}
\end{figure*}

\begin{figure*}
    \centering
    \includegraphics[width=0.85\linewidth]{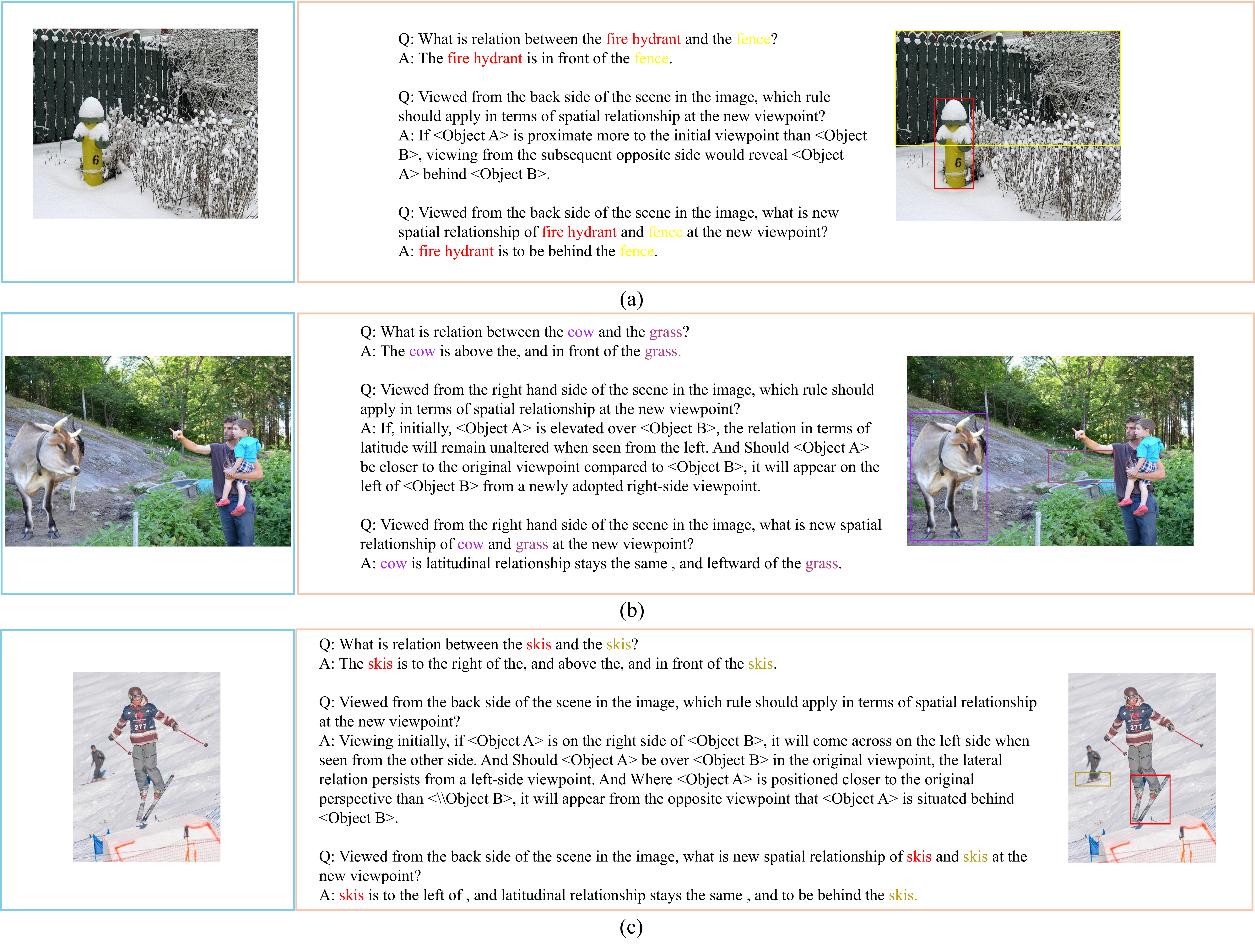}
    \caption{Data Examples of SpaceSGG-Conv in SpaceSGG.}
    \label{fig:supp-conv}
\end{figure*}

We provide three types examples of SpaceSGG data, see in Figure \ref{fig:supp-desc}, Figure \ref{fig:supp-qa} and Figure \ref{fig:supp-conv}.

\section{PSG Dataset Evaluation Comparison}

\begin{figure*}
    \centering
    \includegraphics[width=0.90\linewidth]{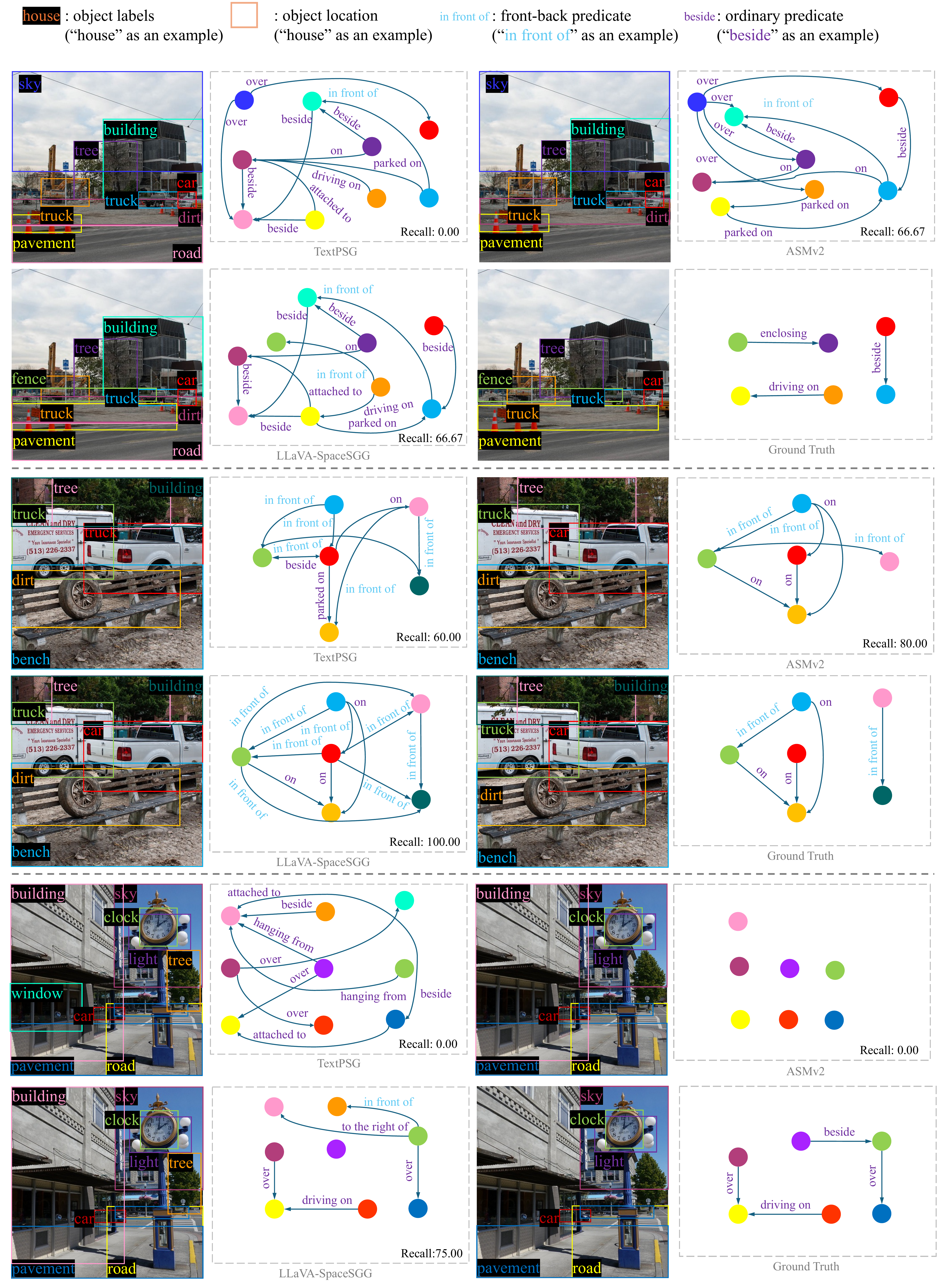}
    \caption{Additional examples of LLaVA-SpaceSGG Open-Vocabulary SGG prediction compared with others on PSG validation set.}
    \label{fig:supp-eval}
\end{figure*}

We report more visual evaluation results of LLaVA-SpaceSGG compared with ASMv2 and TextPSG, see in Figure \ref{fig:supp-eval}.

\vspace{5cm}

\begin{table}[t]
    \centering
    \resizebox{\linewidth}{!}{ 
    \rowcolors{2}{gray!25}{white}  
    \begin{tabular}{p{3cm}ccc}  
        \toprule
        \textbf{Ablation Setting} & \textbf{Recall} & \textbf{mRecall} & \textbf{Accuracy (\%)} \\
        \midrule
        LLaVA-SpaceSGG\newline-ab-data-6 & 9.49& 8.18& 30.415\\
        LLaVA-SpaceSGG\newline-ab-data-7 & \textcolor{blue}{14.95}& \textcolor{blue}{11.74}& \textcolor{green}{51.775}\\
        LLaVA-SpaceSGG\newline-ab-data-8 & \textcolor{green}{13.2}& 8.44& \textcolor{blue}{51.8}\\
        LLaVA-SpaceSGG\newline-ab-data-9 & 0& 0& 0\\
        LLaVA-SpaceSGG\newline-ab-data-10 & 0& 0& 26.895\\
        LLaVA-SpaceSGG\newline-ab-data-11 & 13.07& \textcolor{green}{8.66}& 39.075\\
        LLaVA-SpaceSGG & \textcolor{red}{15.43} & \textcolor{red}{13.23} & \textcolor{blue}{52.48} \\
        \bottomrule
    \end{tabular}
    }
    \caption{We experimented with different mixing ratios of replaced placebo data, using refabricated data combinations for the experimental settings. The \textcolor{red}{red}, \textcolor{blue}{blue}, and \textcolor{green}{green} colors denote the best, the second highest and the third highest results, respectively. For detailed experimental settings, please refer to Figure \ref{fig:placebo-ablations}.}
    \label{tab:placebo-evaluation}
\end{table}

\end{document}